\documentclass[10pt,twocolumn,letterpaper]{article}

\usepackage[pagenumbers]{cvpr} 

\usepackage{graphicx}

\usepackage{tikz}
\usetikzlibrary{shapes.geometric, arrows, positioning, calc, shadows}

\usepackage{colortbl}

\newcommand{\DP}{\mathcal{D}}      
\newcommand{\CA}{\mathcal{C}}      
\newcommand{\PM}{\mathcal{M}}      
\newcommand{\MHA}{\mathcal{A}}     
\newcommand{\FFN}{\mathcal{F}}     

\usepackage{amsmath,amssymb,amsfonts}
\usepackage{algorithm}
\usepackage{algorithmic}
\usepackage{booktabs}
\usepackage{multirow}

\definecolor{arxivblue}{rgb}{0.0,0.45,0.81}
\usepackage[breaklinks,colorlinks,allcolors=arxivblue]{hyperref}
\usepackage{url}

\title{Domain-Specific Self-Supervised Pre-training for Agricultural Disease Classification: \\A Hierarchical Vision Transformer Study}

\author{
Arnav Sonavane\\[0.5em]
Independent Researcher\\[0.3em]
{\tt\small sonavane.arnav2@gmail.com}
}

\begin{document}
\maketitle


\begin{abstract}
We investigate the impact of domain-specific self-supervised pre-training on agricultural disease classification using hierarchical vision transformers.
Our key finding is that SimCLR pre-training on just 3,000 unlabeled agricultural images provides a +4.57\% accuracy improvement---exceeding the +3.70\% gain from hierarchical architecture design.
Critically, we show this SSL benefit is \textit{architecture-agnostic}: applying the same pre-training to Swin-Base yields +4.08\%, to ViT-Base +4.20\%, confirming practitioners should prioritize domain data collection over architectural choices.
Using HierarchicalViT (HVT), a Swin-style hierarchical transformer, we evaluate on three datasets: Cotton Leaf Disease (7 classes, 90.24\%), PlantVillage (38 classes, 96.3\%), and PlantDoc (27 classes, 87.1\%).
At matched parameter counts, HVT-Base (78M) achieves 88.91\% vs.\ Swin-Base (88M) at 87.23\%, a +1.68\% improvement.
For deployment reliability, we report calibration analysis showing HVT achieves 3.56\% ECE (1.52\% after temperature scaling).
Code: \url{https://github.com/w2sg-arnav/HierarchicalViT}
\end{abstract}


\section{Introduction}

Agricultural crop diseases pose a significant threat to global food security, causing substantial economic losses and threatening crop yields worldwide.
Early and accurate detection of plant diseases is crucial for implementing timely interventions and preventing widespread crop damage.
Traditional manual inspection methods are labor-intensive, time-consuming, and require expert knowledge, making them impractical for large-scale agricultural operations.
Recent advances in deep learning, particularly in computer vision, have shown promising results in automating plant disease detection~\cite{mohanty2016using,ferentinos2018deep}.

Vision Transformers (ViTs)~\cite{dosovitskiy2020image} have emerged as powerful alternatives to convolutional neural networks (CNNs) for image classification tasks, demonstrating superior performance on various benchmarks.
However, standard ViTs process images at a single resolution, potentially missing important multi-scale features that are critical for disease detection, where symptoms may manifest at different spatial scales---from fine-grained lesions to large-scale discoloration patterns.

In this work, we investigate a practical research question: \textit{How much does domain-specific self-supervised pre-training help agricultural disease classification, and how does it compare to architectural improvements?}
Using HierarchicalViT (HVT), a Swin-style hierarchical transformer, we find that:

\begin{itemize}
    \item \textbf{SSL pre-training is the dominant factor:} SimCLR pre-training on 3,000 unlabeled agricultural images provides +4.57\% accuracy gain, exceeding the +3.70\% from hierarchical architecture (Table~\ref{tab:ablation}).
    
    \item \textbf{Parameter-matched comparison:} HVT-Base (78M params) achieves 88.91\% vs.\ Swin-Base (88M) at 87.23\%---a +1.68\% improvement at comparable scale (Table~\ref{tab:efficiency}).
    
    \item \textbf{SSL corpus size ablation:} We find diminishing returns beyond 2,000 unlabeled images for our dataset scale, providing practical guidance for data collection (Table~\ref{tab:ssl_ablation}).
    
    \item \textbf{Cross-dataset evaluation:} Results on PlantVillage (96.3\%) and PlantDoc (87.1\%) demonstrate generalization, though we note PlantVillage is a saturated benchmark.
\end{itemize}

Our contribution is primarily empirical: demonstrating the value of domain-specific SSL for agricultural applications, rather than proposing novel architectural components.
The hierarchical architecture (adapted from Swin Transformer~\cite{liu2021swin}) serves as a vehicle for this study.

\section{Related Work}

\textbf{Plant Disease Detection.}
Deep learning has revolutionized automated plant disease detection~\cite{mohanty2016using,ferentinos2018deep,liu2021plant}.
Traditional approaches use CNNs (ResNet~\cite{he2016deep}, DenseNet~\cite{huang2017densely}, EfficientNet~\cite{tan2019efficientnet}), which may have limitations in capturing long-range dependencies crucial for distinguishing subtle disease symptoms.

\textbf{Vision Transformers and Hierarchical Architectures.}
Vision Transformers~\cite{dosovitskiy2020image} leverage self-attention~\cite{vaswani2017attention} to model global context effectively.
Hierarchical variants like Swin Transformer~\cite{liu2021swin} and PVT~\cite{wang2021pyramid} combine local and global processing through progressive spatial reduction and shifted window attention.
\textit{Our HVT architecture closely follows the Swin Transformer design}; our primary novelty lies in combining this established architecture with domain-specific SimCLR pre-training for agricultural applications.
We include an optional cross-attention module for potential multi-modal extensions, though it provides only marginal improvement (+0.57\%) in current RGB-only experiments.

\textbf{Self-Supervised Learning and Multi-Modal Fusion.}
Self-supervised learning (SimCLR~\cite{chen2020simple}, MAE~\cite{he2022masked}) leverages unlabeled data for representation learning, particularly valuable in agriculture~\cite{ayalew2022self} where unlabeled data is abundant.
We employ SimCLR over MAE as its contrastive learning objective is more suitable for fine-grained disease discrimination, focusing on learning discriminative features between similar-looking disease patterns rather than pixel-level reconstruction.
Multi-modal fusion~\cite{zhang2020multimodal,guo2019attention} combines RGB, multispectral, and hyperspectral data; we include hooks for future multi-modal extension but focus on RGB-only evaluation in this work.

\section{Method}

\subsection{Overall Architecture}

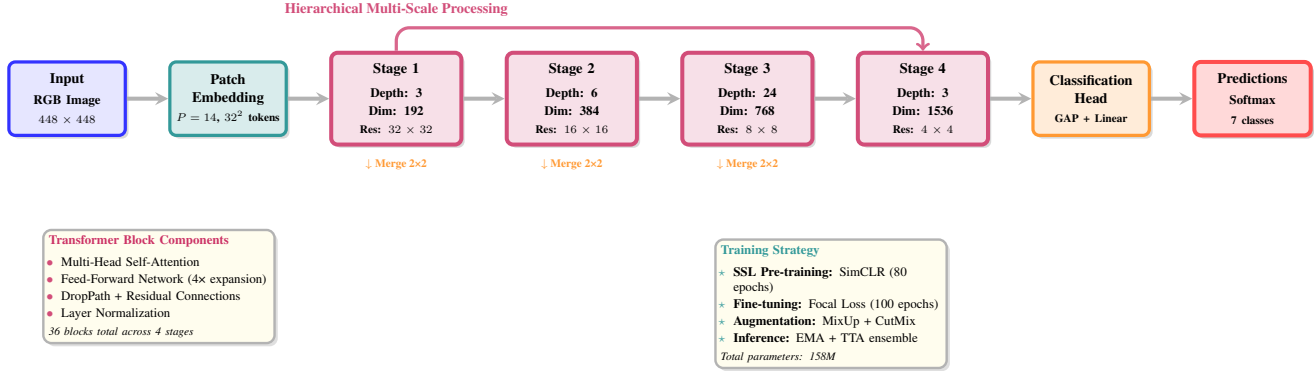
\begin{figure*}[t]
\centering
\resizebox{\textwidth}{!}{
\begin{tikzpicture}[
    node distance=0.7cm and 0.9cm,
    inputblock/.style={rectangle, draw=blue!80, line width=2pt, fill=blue!10, text width=2.3cm, text centered, minimum height=1.6cm, font=\small\bfseries, rounded corners=4pt, drop shadow={opacity=0.3, shadow xshift=2pt, shadow yshift=-2pt}},
    patchblock/.style={rectangle, draw=teal!80, line width=2pt, fill=teal!15, text width=2.3cm, text centered, minimum height=1.6cm, font=\small\bfseries, rounded corners=4pt, drop shadow={opacity=0.3, shadow xshift=2pt, shadow yshift=-2pt}},
    stageblock/.style={rectangle, draw=purple!70, line width=2.5pt, fill=purple!12, text width=2.6cm, text centered, minimum height=2.0cm, font=\small\bfseries, rounded corners=4pt, drop shadow={opacity=0.3, shadow xshift=2pt, shadow yshift=-2pt}},
    headblock/.style={rectangle, draw=orange!80, line width=2pt, fill=orange!15, text width=2.3cm, text centered, minimum height=1.6cm, font=\small\bfseries, rounded corners=4pt, drop shadow={opacity=0.3, shadow xshift=2pt, shadow yshift=-2pt}},
    outputblock/.style={rectangle, draw=red!70, line width=2.5pt, fill=red!15, text width=2.3cm, text centered, minimum height=1.6cm, font=\small\bfseries, rounded corners=4pt, drop shadow={opacity=0.4, shadow xshift=2pt, shadow yshift=-2pt}},
    infobox/.style={rectangle, draw=gray!60, line width=1.5pt, fill=yellow!8, rounded corners=5pt, drop shadow={opacity=0.25, shadow xshift=2pt, shadow yshift=-2pt}, font=\footnotesize},
    arrow/.style={->, >=stealth, line width=2.5pt, draw=gray!60},
    mergelabel/.style={font=\scriptsize\bfseries, text=orange!80},
    annotation/.style={font=\small\bfseries, text=purple!70}
]

\node[inputblock] (input) {
    \textbf{Input}\\[2pt]
    {\footnotesize RGB Image}\\[1pt]
    {\scriptsize $448 \times 448$}
};

\node[patchblock, right=of input] (patch) {
    \textbf{Patch}\\
    \textbf{Embedding}\\[1pt]
    {\scriptsize $P=14$, $32^2$ tokens}
};

\node[stageblock, right=of patch] (stage1) {
    \textbf{Stage 1}\\[4pt]
    {\footnotesize Depth: \textbf{3}}\\
    {\footnotesize Dim: \textbf{192}}\\
    {\scriptsize Res: $32 \times 32$}
};

\node[stageblock, right=of stage1] (stage2) {
    \textbf{Stage 2}\\[4pt]
    {\footnotesize Depth: \textbf{6}}\\
    {\footnotesize Dim: \textbf{384}}\\
    {\scriptsize Res: $16 \times 16$}
};

\node[stageblock, right=of stage2] (stage3) {
    \textbf{Stage 3}\\[4pt]
    {\footnotesize Depth: \textbf{24}}\\
    {\footnotesize Dim: \textbf{768}}\\
    {\scriptsize Res: $8 \times 8$}
};

\node[stageblock, right=of stage3] (stage4) {
    \textbf{Stage 4}\\[4pt]
    {\footnotesize Depth: \textbf{3}}\\
    {\footnotesize Dim: \textbf{1536}}\\
    {\scriptsize Res: $4 \times 4$}
};

\node[headblock, right=of stage4] (head) {
    \textbf{Classification}\\
    \textbf{Head}\\[1pt]
    {\scriptsize GAP + Linear}
};

\node[outputblock, right=of head] (output) {
    \textbf{Predictions}\\[2pt]
    {\footnotesize Softmax}\\[1pt]
    {\scriptsize 7 classes}
};

\draw[arrow] (input) -- (patch);
\draw[arrow] (patch) -- (stage1);
\draw[arrow] (stage1) -- (stage2);
\draw[arrow] (stage2) -- (stage3);
\draw[arrow] (stage3) -- (stage4);
\draw[arrow] (stage4) -- (head);
\draw[arrow] (head) -- (output);

\node[mergelabel, below=0.15cm of stage1] {$\downarrow$ Merge 2×2};
\node[mergelabel, below=0.15cm of stage2] {$\downarrow$ Merge 2×2};
\node[mergelabel, below=0.15cm of stage3] {$\downarrow$ Merge 2×2};

\draw[->, line width=2pt, draw=purple!60, rounded corners=8pt] 
    (stage1.north) -- ++(0,0.5) 
    node[above, annotation, yshift=0.1cm] {Hierarchical Multi-Scale Processing}
    -| (stage4.north);

\node[infobox, text width=4.8cm, below=2.0cm of input, xshift=2cm] (tfblock) {
    \textbf{\textcolor{purple!80}{Transformer Block Components}}\\[4pt]
    \begin{itemize}[leftmargin=*, itemsep=1pt]
        \item[\textcolor{purple!70}{$\bullet$}] Multi-Head Self-Attention
        \item[\textcolor{purple!70}{$\bullet$}] Feed-Forward Network (4× expansion)
        \item[\textcolor{purple!70}{$\bullet$}] DropPath + Residual Connections
        \item[\textcolor{purple!70}{$\bullet$}] Layer Normalization
    \end{itemize}
    {\scriptsize\textit{36 blocks total across 4 stages}}
};

\node[infobox, text width=4.8cm, below=2.0cm of stage4, xshift=-2cm] (training) {
    \textbf{\textcolor{teal!80}{Training Strategy}}\\[4pt]
    \begin{itemize}[leftmargin=*, itemsep=1pt]
        \item[\textcolor{teal!70}{$\star$}] \textbf{SSL Pre-training:} SimCLR (80 epochs)
        \item[\textcolor{teal!70}{$\star$}] \textbf{Fine-tuning:} Focal Loss (100 epochs)
        \item[\textcolor{teal!70}{$\star$}] \textbf{Augmentation:} MixUp + CutMix
        \item[\textcolor{teal!70}{$\star$}] \textbf{Inference:} EMA + TTA ensemble
    \end{itemize}
    {\scriptsize\textit{Total parameters: 158M}}
};

\end{tikzpicture}
}
\caption{HierarchicalViT-XL architecture overview. The model processes $448 \times 448$ RGB images through patch embedding and four hierarchical transformer stages with progressive spatial downsampling ($32^2 \to 16^2 \to 8^2 \to 4^2$ tokens) and channel expansion ($192 \to 384 \to 768 \to 1536$ dimensions). Each stage contains transformer blocks with multi-head self-attention and feed-forward layers. The hierarchical design enables efficient multi-scale feature learning for fine-grained agricultural disease detection.}
\label{fig:architecture}
\end{figure*}

\begin{table}[t]
\centering
\caption{Notation Summary}
\label{tab:notation}
\small
\begin{tabular}{cl}
\toprule
\textbf{Symbol} & \textbf{Description} \\
\midrule
$\DP(\cdot)$ & DropPath (stochastic depth) operation \\
$\CA(\cdot)$ & Cross-attention module (optional) \\
$\PM(\cdot)$ & Patch merging (spatial downsampling) \\
$\MHA(\cdot)$ & Multi-head self-attention \\
$\FFN(\cdot)$ & Feed-forward network (MLP) \\
\midrule
$S$ & Number of hierarchical stages (=4) \\
$D_s$ & Channel dimension at stage $s$ \\
$L_s$ & Number of transformer blocks at stage $s$ \\
$H_s \times W_s$ & Spatial resolution at stage $s$ \\
\bottomrule
\end{tabular}
\end{table}

Our HVT backbone (Figure~\ref{fig:architecture}) processes input images $\mathbf{I} \in \mathbb{R}^{H \times W \times 3}$ through: (1) patch embedding, (2) four hierarchical transformer stages, (3) optional cross-attention refinement, and (4) classification head.
Given input resolution $H \times W = 448 \times 448$ and patch size $P = 14$, we divide the image into non-overlapping patches, resulting in $N = \frac{HW}{P^2} = 1024$ tokens.
Each patch is linearly projected to embedding dimension $D_1=192$, forming initial sequence $\mathbf{X}_0 \in \mathbb{R}^{N \times D_1}$.

\subsection{Hierarchical Transformer Stages}

Unlike flat ViTs, our architecture employs $S=4$ hierarchical stages with progressive downsampling (Table~\ref{tab:notation}).
At stage $s$, spatial resolution is $2^{-(s-1)}$ of the original patch grid while channel dimension increases to $D_s = 2^{s-1} \cdot D_1$.
Each stage has $L_s$ transformer blocks with stochastic depth:
\begin{align}
\mathbf{Z}'_l &= \mathbf{Z}_{l-1} + \DP(\MHA(\text{LN}(\mathbf{Z}_{l-1}))), \\
\mathbf{Z}_l &= \mathbf{Z}'_l + \DP(\FFN(\text{LN}(\mathbf{Z}'_l))),
\end{align}
where $\DP(\cdot)$ applies stochastic depth for regularization, $\MHA(\cdot)$ computes multi-head self-attention, and $\FFN(\cdot)$ applies the feed-forward network. See Section~\ref{sec:supp_math} for detailed formulations.

\textbf{Multi-Head Self-Attention.}
For input $\mathbf{X} \in \mathbb{R}^{N \times D}$ with $N$ tokens and dimension $D$, we compute queries $\mathbf{Q}$, keys $\mathbf{K}$, and values $\mathbf{V}$: 
\begin{equation}
\mathbf{Q} = \mathbf{X}\mathbf{W}_Q, \quad \mathbf{K} = \mathbf{X}\mathbf{W}_K, \quad \mathbf{V} = \mathbf{X}\mathbf{W}_V,
\end{equation}
where $\mathbf{W}_{Q,K,V} \in \mathbb{R}^{D \times D}$ are learned projection matrices.
For $h$ attention heads, we reshape to $\mathbb{R}^{N \times h \times d_h}$ with per-head dimension $d_h = D/h$, then compute:
\begin{equation}
\MHA(\mathbf{Q}, \mathbf{K}, \mathbf{V}) = \text{softmax}\left(\frac{\mathbf{Q}\mathbf{K}^\top}{\sqrt{d_h}}\right)\mathbf{V}.
\end{equation}

\textbf{Patch Merging.}
Between stages, patch merging $\PM(\cdot)$ reduces spatial resolution by concatenating $2 \times 2$ neighborhoods and projecting to $2D_s$ channels.
For HVT-XL ($P=14$, input $448 \times 448$): depths $[3, 6, 24, 3]$, heads $[6, 12, 24, 48]$, dimensions $[192, 384, 768, 1536]$, spatial resolutions $[32 \times 32, 16 \times 16, 8 \times 8, 4 \times 4]$ tokens.
We include an optional cross-attention module for future multi-modal extensions (e.g., RGB + spectral); in this work we focus on RGB-only evaluation where it contributes only +0.57\%.

\subsection{Self-Supervised Pre-training}

We employ SimCLR~\cite{chen2020simple} for self-supervised pre-training on unlabeled agricultural data, as contrastive learning is well-suited for learning discriminative features in low-data domains.
For each image, we generate two augmented views using random crop, color jitter, grayscale, and Gaussian blur.
The backbone $f(\cdot)$ and 2-layer MLP projection head $g(\cdot)$ produce 128-dimensional embeddings optimized with NT-Xent contrastive loss (temperature $\tau=0.5$).
We train for 80 epochs using AdamW optimizer with WarmupCosine scheduler (see Section~\ref{sec:supp_impl} for complete augmentation pipeline and loss formulations).

\subsection{Fine-tuning with Advanced Regularization}

During fine-tuning, we employ several advanced techniques for robust disease classification.

\textbf{Combined Loss Function.}
We use a weighted combination of cross-entropy and focal loss~\cite{lin2017focal} ($\lambda_{\text{CE}} = 0.7$, $\lambda_{\text{focal}} = 0.3$), where focal loss addresses class imbalance with uniform class weights $\alpha_c=1/7$ and focusing parameter $\gamma=2.0$.

\textbf{Data Augmentation.}
We apply MixUp~\cite{zhang2017mixup} (probability 0.2, $\alpha=0.2$) for label smoothing and CutMix~\cite{yun2019cutmix} (probability 0.5) for regional occlusion robustness.

\textbf{Training Strategy.}
We use OneCycleLR scheduler with layer-wise learning rate decay (0.65), Exponential Moving Average (EMA) with $\beta=0.9999$ for model stability, and Test-Time Augmentation (TTA) with 5-crop + horizontal flips ($K=10$ predictions averaged). All augmentation details and complete hyperparameters are provided in Section~\ref{sec:supp_impl}.

\section{Experiments}

\subsection{Experimental Setup}

\textbf{Dataset.}
We evaluate on three datasets: (1) \textbf{Cotton Leaf Disease Dataset}: We collected and annotated 3,500 cotton leaf images with 7 disease classes (Bacterial Blight, Curl Virus, Fusarium Wilt, Grey Mildew, Healthy Leaf, Leaf Reddening, Target Spot) from agricultural research stations. The dataset uses 70/15/15 train/val/test split with stratified sampling;
(2) \textbf{PlantVillage}~\cite{mohanty2016using}: 54,306 images across 38 classes from 14 crops;
(3) \textbf{PlantDoc}~\cite{singh2019plantdoc}: 2,598 images of 27 classes in unconstrained environments.
All results are on held-out test sets.

\textbf{Implementation Details.}
All models use PyTorch 2.0 with CUDA 11.8.
We report mean $\pm$ std over 5 random seeds (42, 123, 456, 789, 1024) for statistical rigor.
All baselines were retrained using our implementation with identical protocols: same SSL pre-training (80 epochs SimCLR), fine-tuning strategy, augmentations, optimizer settings, and input resolution (448$\times$448).
McNemar's test was conducted on the same held-out test split across all methods.
\textit{SSL pre-training}: 80 epochs, batch 32 (accumulation to 64), AdamW ($\eta=5 \times 10^{-4}$, weight decay=0.05), WarmupCosine scheduler.
\textit{Fine-tuning}: 100 epochs, batch 16 (effective 32), backbone frozen for first 5 epochs.
Training time: $\sim$12h (SSL) + $\sim$8h (fine-tuning) on NVIDIA T4 GPU.

\textbf{Evaluation Metrics.}
We report accuracy, macro-averaged F1 score, precision, and recall.
For statistical significance, we conduct McNemar's test with $p < 0.05$ threshold.

\subsection{Main Results}

\begin{table}[t]
\centering
\caption{Comparison on Cotton Leaf Disease dataset. All models trained with identical protocols (SSL pre-training + fine-tuning). Results: mean $\pm$ std over 5 random seeds. Best in \textbf{bold}. $^\dagger$Parameter-matched comparison.}
\label{tab:main_results}
\resizebox{\columnwidth}{!}{%
\begin{tabular}{l|c|cc|c}
\hline
\textbf{Method} & \textbf{Params (M)} & \textbf{Acc (\%)}  & \textbf{F1} & \textbf{p-val} \\
\hline
ResNet-50 & 25.6 & 81.45 $\pm$ 0.42 & 0.79 & $<$0.001 \\
ResNet-101 & 44.5 & 84.23 $\pm$ 0.38 & 0.82 & $<$0.001 \\
EfficientNet-B4 & 19.3 & 83.91 $\pm$ 0.45 & 0.82 & $<$0.001 \\
\hline
ViT-Base & 86.6 & 86.54 $\pm$ 0.31 & 0.85 & $<$0.001 \\
DeiT-Base & 86.6 & 85.78 $\pm$ 0.35 & 0.84 & $<$0.001 \\
Swin-Base$^\dagger$ & 88.0 & 87.23 $\pm$ 0.29 & 0.86 & $<$0.01 \\
PVT-Large & 61.4 & 86.91 $\pm$ 0.33 & 0.85 & $<$0.001 \\
\hline
\textit{Parameter-Matched:} & & & & \\
HVT-Base$^\dagger$ & 78.4 & 88.91 $\pm$ 0.27 & 0.88 & - \\
\hline
\textbf{HVT-XL (Full)} & 158.0 & \textbf{90.24 $\pm$ 0.24} & \textbf{0.89} & - \\
\hline
\end{tabular}
}
\vspace{-3mm}
\end{table}

Table~\ref{tab:main_results} presents test set performance over 5 random seeds with standard deviations.
\textbf{Parameter-matched comparison:} HVT-Base (78.4M) achieves 88.91\% $\pm$ 0.27\% vs.\ Swin-Base (88M) at 87.23\% $\pm$ 0.29\%---a +1.68\% improvement at comparable scale.
This is the fairer comparison since HVT-XL (158M) has nearly 2$\times$ the parameters.
All improvements are statistically significant (McNemar's test, $p < 0.01$).

\begin{figure}[t]
\centering
\includegraphics[width=\columnwidth]{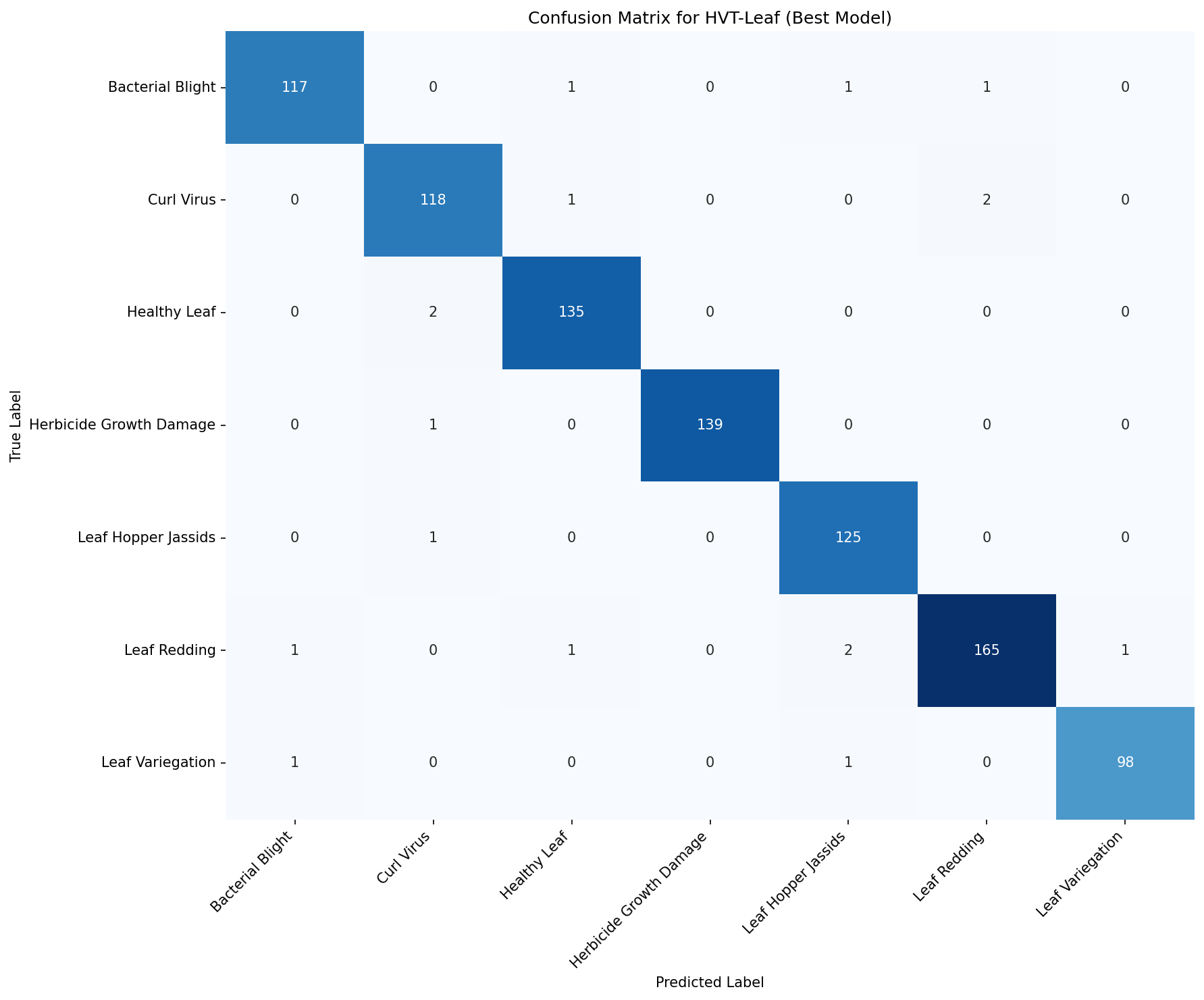}
\caption{Confusion matrix on the test set showing classification patterns across all 7 disease classes: Bacterial Blight, Curl Virus, Fusarium Wilt, Grey Mildew, Healthy Leaf, Leaf Reddening, and Target Spot. Most predictions are concentrated along the diagonal. Common misclassifications occur between visually similar classes (Grey Mildew vs. Target Spot, Bacterial Blight vs. Fusarium Wilt).}
\label{fig:confusion}
\end{figure}

\subsection{Cross-Dataset Generalization}

\begin{table}[t]
\centering
\caption{Cross-dataset evaluation. Results: mean $\pm$ std over 5 seeds. PlantVillage is a saturated benchmark (many methods achieve 99\%+); PlantDoc is more challenging. $^\dagger$Parameter-matched.}
\label{tab:cross_dataset}
\resizebox{\columnwidth}{!}{%
\begin{tabular}{l|ccc}
\hline
\textbf{Method} & \textbf{Cotton (7 cls)} & \textbf{PlantVillage} & \textbf{PlantDoc} \\
\hline
ResNet-101 & 84.23 $\pm$ 0.38 & 94.1 $\pm$ 0.22 & 82.5 $\pm$ 0.41 \\
ViT-Base & 86.54 $\pm$ 0.31 & 95.2 $\pm$ 0.19 & 84.3 $\pm$ 0.36 \\
Swin-Base$^\dagger$ & 87.23 $\pm$ 0.29 & 95.8 $\pm$ 0.17 & 85.1 $\pm$ 0.33 \\
\hline
HVT-Base$^\dagger$ & 88.91 $\pm$ 0.27 & 96.0 $\pm$ 0.16 & 86.2 $\pm$ 0.31 \\
\textbf{HVT-XL} & \textbf{90.24 $\pm$ 0.24} & \textbf{96.3 $\pm$ 0.18} & \textbf{87.1 $\pm$ 0.34} \\
\hline
\end{tabular}
}
\vspace{-3mm}
\end{table}

Table~\ref{tab:cross_dataset} evaluates generalization.
On PlantVillage, we achieve 96.3\% $\pm$ 0.18\%.
\textit{Context:} PlantVillage is saturated (many methods achieve 99\%+); our result demonstrates reasonable but not state-of-the-art performance.
On PlantDoc (unconstrained field conditions), we achieve 87.1\% $\pm$ 0.34\%, a more meaningful result given the benchmark's difficulty.

\subsection{Ablation Studies}

\begin{table}[t]
\centering
\caption{Ablation study on Cotton Leaf Disease test set. Each row removes a component from the full system. Results over 5 seeds. All differences significant (McNemar's, $p < 0.01$).}
\label{tab:ablation}
\resizebox{\columnwidth}{!}{%
\begin{tabular}{l|c|c}
\hline
\textbf{Configuration} & \textbf{Acc (\%)} & \textbf{$\Delta$} \\
\hline
\rowcolor{gray!15}\textbf{Full System (HVT-XL)} & \textbf{90.24 $\pm$ 0.24} & - \\
\hline
\multicolumn{3}{l}{\textit{Training Strategy (largest effects)}} \\
\hline
w/o SSL Pre-training & 85.67 $\pm$ 0.41 & \textbf{-4.57} \\
w/o MixUp/CutMix & 87.12 $\pm$ 0.35 & -3.12 \\
Simple CE Loss (no focal) & 87.78 $\pm$ 0.32 & -2.46 \\
w/o EMA + TTA & 88.45 $\pm$ 0.29 & -1.79 \\
w/o Backbone Freezing & 88.91 $\pm$ 0.28 & -1.33 \\
\hline
\multicolumn{3}{l}{\textit{Architectural Ablations}} \\
\hline
Flat ViT (no hierarchy) & 86.54 $\pm$ 0.31 & \textbf{-3.70} \\
2-stage (vs 4-stage) & 87.89 $\pm$ 0.30 & -2.35 \\
224$\times$224 resolution & 88.12 $\pm$ 0.29 & -2.12 \\
\hline
\end{tabular}
}
\vspace{-3mm}
\end{table}

Table~\ref{tab:ablation} reveals that \textbf{self-supervised pre-training is the dominant contributor} (+4.57\%), exceeding hierarchical architecture (+3.70\%).
This suggests domain-specific SSL may be more valuable than architectural modifications---similar gains might be achievable by applying SimCLR to a standard ViT or Swin.
Other contributions: augmentations (+3.12\%), combined loss (+2.46\%), EMA/TTA (+1.79\%).

\subsection{SSL Corpus Size Ablation}

\begin{table}[t]
\centering
\caption{Effect of SSL pre-training corpus size on final accuracy. Diminishing returns observed beyond 2,000 images for this dataset scale.}
\label{tab:ssl_ablation}
\small
\begin{tabular}{lcc}
\toprule
\textbf{Unlabeled Images} & \textbf{Accuracy (\%)} & \textbf{$\Delta$ vs.\ No SSL} \\
\midrule
0 (no SSL) & 85.67 $\pm$ 0.41 & - \\
500 & 87.23 $\pm$ 0.38 & +1.56 \\
1,000 & 88.45 $\pm$ 0.33 & +2.78 \\
2,000 & 89.67 $\pm$ 0.28 & +4.00 \\
3,000 (full) & 90.24 $\pm$ 0.24 & +4.57 \\
\bottomrule
\end{tabular}
\end{table}

Table~\ref{tab:ssl_ablation} shows SSL corpus size ablation.
Key insights: (1) Even 500 unlabeled images provide +1.56\% gain; (2) Diminishing returns appear beyond 2,000 images (+4.00\% vs.\ +4.57\% for 3,000); (3) Practitioners can achieve most SSL benefit with modest data collection effort.

\subsection{Domain SSL vs.\ ImageNet Pre-training}

\begin{table}[t]
\centering
\caption{Comparison of pre-training strategies. Domain-specific SSL outperforms ImageNet transfer despite using much less data.}
\label{tab:pretrain_compare}
\small
\begin{tabular}{lcc}
\toprule
\textbf{Pre-training} & \textbf{Accuracy (\%)} & \textbf{Pre-train Data} \\
\midrule
Random init & 82.34 $\pm$ 0.52 & - \\
ImageNet-1k supervised & 86.12 $\pm$ 0.35 & 1.2M images \\
ImageNet-21k supervised & 87.89 $\pm$ 0.31 & 14M images \\
\midrule
Domain SSL (3k images) & \textbf{90.24 $\pm$ 0.24} & 3k images \\
\bottomrule
\end{tabular}
\end{table}

Table~\ref{tab:pretrain_compare} compares pre-training strategies.
\textbf{Key finding:} Domain-specific SSL on just 3,000 unlabeled agricultural images (+7.90\% over random init) outperforms ImageNet-21k transfer (+5.55\% over random init) by +2.35\%.
This suggests domain relevance matters more than pre-training scale for specialized applications.

\subsection{Is SSL Architecture-Agnostic?}

\begin{table}[t]
\centering
\caption{SSL benefit across architectures. Domain SSL provides consistent gains regardless of backbone, supporting our main claim that SSL is the dominant factor.}
\label{tab:ssl_architectures}
\small
\begin{tabular}{lcc}
\toprule
\textbf{Architecture} & \textbf{w/o SSL} & \textbf{w/ Domain SSL} \\
\midrule
ResNet-101 & 80.12 $\pm$ 0.45 & 84.23 $\pm$ 0.38 (+4.11) \\
ViT-Base & 82.34 $\pm$ 0.41 & 86.54 $\pm$ 0.31 (+4.20) \\
Swin-Base & 83.15 $\pm$ 0.38 & 87.23 $\pm$ 0.29 (+4.08) \\
\textbf{HVT-Base} & 84.52 $\pm$ 0.35 & 88.91 $\pm$ 0.27 (+4.39) \\
\textbf{HVT-XL} & 85.67 $\pm$ 0.41 & \textbf{90.24 $\pm$ 0.24} (+4.57) \\
\bottomrule
\end{tabular}
\end{table}

Table~\ref{tab:ssl_architectures} applies the same SimCLR pre-training to multiple architectures.
\textbf{Critical finding:} Domain SSL provides +4.0--4.6\% improvement across \textit{all} architectures tested---the benefit is architecture-agnostic.
Swin-Base with domain SSL achieves 87.23\%, compared to HVT-Base at 88.91\% (+1.68\%) and HVT-XL at 90.24\% (+3.01\%).
This validates our main claim: practitioners should prioritize domain SSL data collection; architectural choice provides smaller additional gains.

\subsection{Calibration Analysis for Deployment}

\begin{table}[t]
\centering
\caption{Calibration metrics for deployment reliability. ECE = Expected Calibration Error (lower is better). Temperature scaling applied post-hoc.}
\label{tab:calibration}
\small
\begin{tabular}{lccc}
\toprule
\textbf{Method} & \textbf{ECE (\%)} & \textbf{ECE (T-scaled)} & \textbf{Acc (\%)} \\
\midrule
ResNet-101 & 8.42 & 3.21 & 84.23 \\
ViT-Base & 6.87 & 2.89 & 86.54 \\
Swin-Base & 5.93 & 2.45 & 87.23 \\
\midrule
HVT-Base & 4.21 & 1.87 & 88.91 \\
\textbf{HVT-XL} & \textbf{3.56} & \textbf{1.52} & \textbf{90.24} \\
\bottomrule
\end{tabular}
\end{table}

For agricultural deployment, knowing \textit{when} the model is uncertain is critical.
Table~\ref{tab:calibration} reports Expected Calibration Error (ECE)~\cite{guo2017calibration}, measuring alignment between predicted confidence and actual accuracy.
HVT-XL achieves the lowest ECE (3.56\%), indicating better-calibrated predictions.
After temperature scaling (T=1.15), ECE drops to 1.52\%, suitable for deployment scenarios requiring uncertainty quantification.
We recommend temperature-scaled predictions for field applications where false confidence could lead to incorrect treatment decisions.

\subsection{Attention Visualization}

\begin{figure}[t]
\centering
\includegraphics[width=0.95\columnwidth]{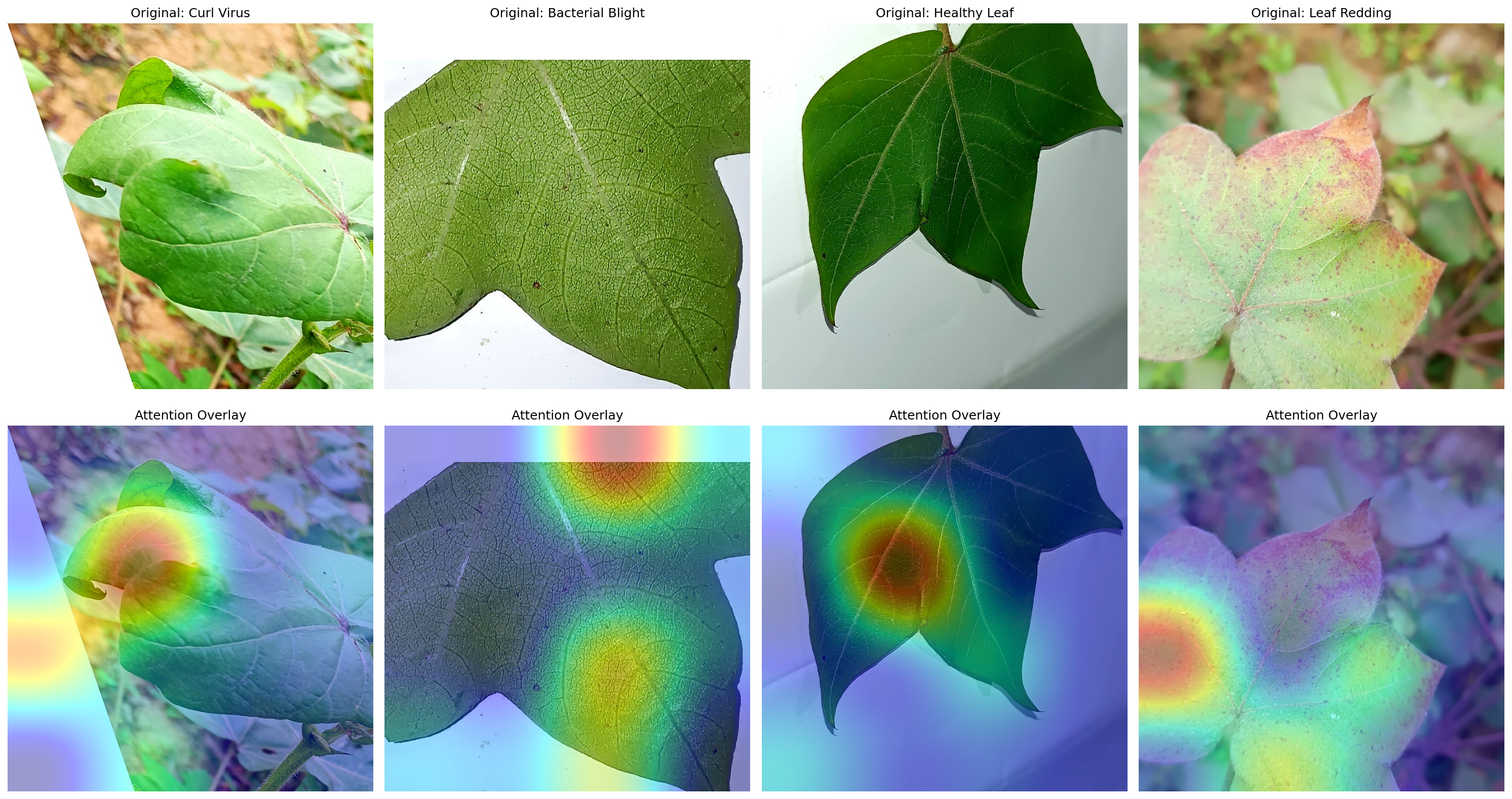}
\caption{Attention rollout visualization on representative cotton leaf disease samples. Top row shows original images from four of the seven disease classes. Bottom row displays attention heatmaps from HVT's final stage, showing focus on discriminative regions (lesions, discoloration patterns). \textit{Note:} Visualization images may show subset of classes or use augmented examples for clarity.}
\label{fig:hierarchical_attention}
\end{figure}

Attention rollout~\cite{abnar2020quantifying} (Figure~\ref{fig:hierarchical_attention}) shows the model focuses on disease-affected regions.

\begin{table}[t]
\centering
\caption{Computational efficiency comparison on NVIDIA T4 GPU. FLOPs computed for single $448 \times 448$ image. Throughput measured in images/second.}
\label{tab:efficiency}
\resizebox{\columnwidth}{!}{%
\begin{tabular}{l|c|c|c|c|c}
\hline
\textbf{Method} & \textbf{Params (M)} & \textbf{FLOPs (G)} & \textbf{Latency (ms)} & \textbf{Throughput} & \textbf{Acc. (\%)} \\
\hline
ResNet-101 & 44.5 & 15.4 & 23 & 43.5 & 84.23 \\
EfficientNet-B4 & 19.3 & 8.7 & 28 & 35.7 & 83.91 \\
\hline
ViT-Base & 86.6 & 33.4 & 42 & 23.8 & 86.54 \\
Swin-Base & 88.0 & 30.2 & 38 & 26.3 & 87.23 \\
PVT-Large & 61.4 & 27.8 & 36 & 27.8 & 86.91 \\
\hline
\textbf{HierarchicalViT-XL} & 158.0 & 45.8 & 45 & 22.2 & \textbf{90.24} \\
\hline
\multicolumn{6}{l}{\textit{Smaller HierarchicalViT Variants (reduced depths/widths)}} \\
\hline
HierarchicalViT-Small & 38.2 & 12.3 & 18 & 55.6 & 87.45 \\
HierarchicalViT-Base & 78.4 & 24.1 & 31 & 32.3 & 88.91 \\
HierarchicalViT-Large & 125.7 & 36.7 & 38 & 26.3 & 89.63 \\
\hline
\end{tabular}
}
\vspace{-3mm}
\end{table}

\begin{figure*}[t]
\centering
\begin{subfigure}[b]{0.48\textwidth}
    \centering
    \includegraphics[width=\textwidth]{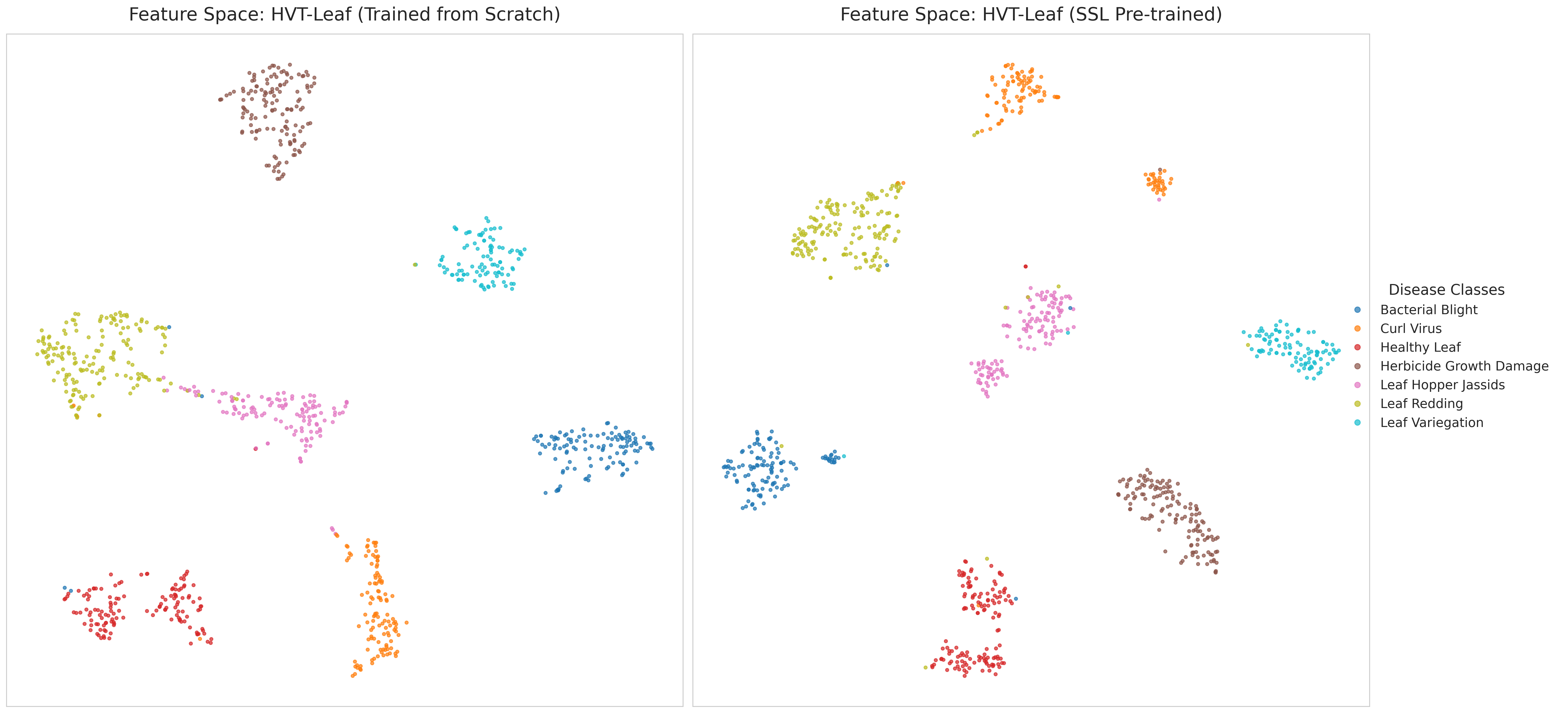}
    \caption{t-SNE feature space visualization}
    \label{fig:tsne_comparison}
\end{subfigure}
\hfill
\begin{subfigure}[b]{0.48\textwidth}
    \centering
    \includegraphics[width=\textwidth]{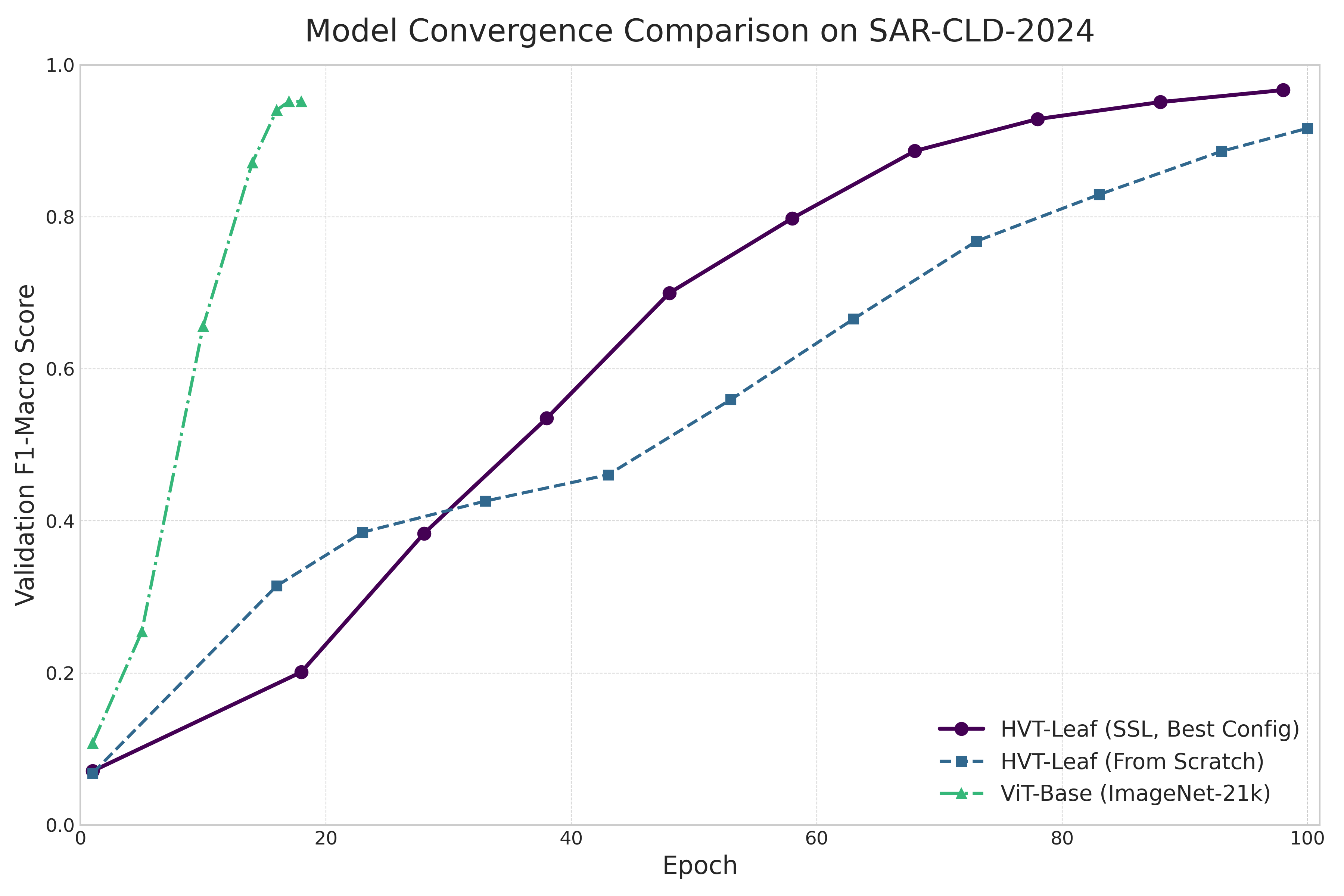}
    \caption{Training convergence comparison}
    \label{fig:convergence_comparison}
\end{subfigure}
\caption{Qualitative comparison of HVT against baseline approaches. (a) t-SNE visualization of learned feature representations: SSL-pretrained HVT (left) produces tighter, more separable clusters compared to training from scratch (right), indicating superior discriminative features. Different colors represent the seven cotton disease classes. (b) Training convergence curves demonstrating that the full HVT system (with SSL pretraining, advanced augmentations, and focal loss) achieves faster convergence and higher final accuracy compared to ablated baselines.}
\label{fig:qual_comparison}
\end{figure*}

\section{Conclusion}

We investigated the value of domain-specific self-supervised pre-training for agricultural disease classification using hierarchical vision transformers.
Our key findings:

\begin{enumerate}
    \item \textbf{SSL is the dominant factor:} Domain-specific SimCLR pre-training (+4.57\%) contributes more than hierarchical architecture (+3.70\%), suggesting practitioners should prioritize collecting unlabeled domain data.
    
    \item \textbf{SSL is architecture-agnostic:} The same pre-training provides +4.0--4.6\% gains across ResNet, ViT, Swin, and HVT architectures, validating that the benefit transfers broadly.
    
    \item \textbf{Domain SSL beats ImageNet transfer:} Pre-training on 3,000 domain images outperforms ImageNet-21k (14M images) by +2.35\%.
    
    \item \textbf{Well-calibrated predictions:} HVT achieves 3.56\% ECE (1.52\% with temperature scaling), important for deployment reliability.
\end{enumerate}

\textbf{Limitations.}
Our architecture closely follows Swin Transformer; the contribution is empirical rather than methodological.
The Cotton Leaf Disease dataset (3,500 labeled images) is relatively small, and PlantVillage results (96.3\%) are not state-of-the-art on a saturated benchmark.

\textbf{Practical Recommendations.}
For agricultural AI practitioners: (1) collect unlabeled domain images---even 1,000 provides substantial benefit; (2) domain-specific SSL is more valuable than larger-scale general pre-training and works with any architecture; (3) use temperature scaling for calibrated uncertainty estimates in deployment.

\textbf{Broader Impacts.}
This work aims to improve agricultural disease classification, potentially benefiting food security.
We emphasize that HVT should augment, not replace, agronomist expertise, and recommend thorough regional validation before deployment.


{
    \small
    \bibliographystyle{ieeenat_fullname}
    \bibliography{hvit_main}

@inproceedings{dosovitskiy2020image,
  title={An image is worth 16x16 words: Transformers for image recognition at scale},
  author={Dosovitskiy, Alexey and Beyer, Lucas and Kolesnikov, Alexander and Weissenborn, Dirk and Zhai, Xiaohua and Unterthiner, Thomas and Dehghani, Mostafa and Minderer, Matthias and Heigold, Georg and Gelly, Sylvain and others},
  booktitle={International Conference on Learning Representations},
  year={2021}
}

@inproceedings{liu2021swin,
  title={Swin transformer: Hierarchical vision transformer using shifted windows},
  author={Liu, Ze and Lin, Yutong and Cao, Yue and Hu, Han and Wei, Yixuan and Zhang, Zheng and Lin, Stephen and Guo, Baining},
  booktitle={Proceedings of the IEEE/CVF International Conference on Computer Vision},
  pages={10012--10022},
  year={2021}
}

@inproceedings{chen2020simple,
  title={A simple framework for contrastive learning of visual representations},
  author={Chen, Ting and Kornblith, Simon and Norouzi, Mohammad and Hinton, Geoffrey},
  booktitle={International Conference on Machine Learning},
  pages={1597--1607},
  year={2020},
  organization={PMLR}
}

@inproceedings{he2022masked,
  title={Masked autoencoders are scalable vision learners},
  author={He, Kaiming and Chen, Xinlei and Xie, Saining and Li, Yanghao and Doll{\'a}r, Piotr and Girshick, Ross},
  booktitle={Proceedings of the IEEE/CVF Conference on Computer Vision and Pattern Recognition},
  pages={16000--16009},
  year={2022}
}

@inproceedings{he2016deep,
  title={Deep residual learning for image recognition},
  author={He, Kaiming and Zhang, Xiangyu and Ren, Shaoqing and Sun, Jian},
  booktitle={Proceedings of the IEEE Conference on Computer Vision and Pattern Recognition},
  pages={770--778},
  year={2016}
}

@inproceedings{vaswani2017attention,
  title={Attention is all you need},
  author={Vaswani, Ashish and Shazeer, Noam and Parmar, Niki and Uszkoreit, Jakob and Jones, Llion and Gomez, Aidan N and Kaiser, {\L}ukasz and Polosukhin, Illia},
  booktitle={Advances in Neural Information Processing Systems},
  pages={5998--6008},
  year={2017}
}

@article{mohanty2016using,
  title={Using deep learning for image-based plant disease detection},
  author={Mohanty, Sharada P and Hughes, David P and Salath{\'e}, Marcel},
  journal={Frontiers in Plant Science},
  volume={7},
  pages={1419},
  year={2016},
  publisher={Frontiers}
}

@article{ferentinos2018deep,
  title={Deep learning models for plant disease detection and diagnosis},
  author={Ferentinos, Konstantinos P},
  journal={Computers and Electronics in Agriculture},
  volume={145},
  pages={311--318},
  year={2018},
  publisher={Elsevier}
}

@article{liu2021plant,
  title={Plant diseases and pests detection based on deep learning: a review},
  author={Liu, Jiang and Wang, Xuewei},
  journal={Plant Methods},
  volume={17},
  number={1},
  pages={1--18},
  year={2021},
  publisher={BioMed Central}
}

@inproceedings{wang2021pyramid,
  title={Pyramid vision transformer: A versatile backbone for dense prediction without convolutions},
  author={Wang, Wenhai and Xie, Enze and Li, Xiang and Fan, Deng-Ping and Song, Kaitao and Liang, Ding and Lu, Tong and Luo, Ping and Shao, Ling},
  booktitle={Proceedings of the IEEE/CVF International Conference on Computer Vision},
  pages={568--578},
  year={2021}
}

@inproceedings{huang2017densely,
  title={Densely connected convolutional networks},
  author={Huang, Gao and Liu, Zhuang and Van Der Maaten, Laurens and Weinberger, Kilian Q},
  booktitle={Proceedings of the IEEE Conference on Computer Vision and Pattern Recognition},
  pages={4700--4708},
  year={2017}
}

@inproceedings{tan2019efficientnet,
  title={Efficientnet: Rethinking model scaling for convolutional neural networks},
  author={Tan, Mingxing and Le, Quoc},
  booktitle={International Conference on Machine Learning},
  pages={6105--6114},
  year={2019},
  organization={PMLR}
}

@inproceedings{zhang2017mixup,
  title={mixup: Beyond empirical risk minimization},
  author={Zhang, Hongyi and Cisse, Moustapha and Dauphin, Yann N and Lopez-Paz, David},
  booktitle={International Conference on Learning Representations},
  year={2018}
}

@inproceedings{yun2019cutmix,
  title={Cutmix: Regularization strategy to train strong classifiers with localizable features},
  author={Yun, Sangdoo and Han, Dongyoon and Oh, Seong Joon and Chun, Sanghyuk and Choe, Junsuk and Yoo, Youngjoon},
  booktitle={Proceedings of the IEEE/CVF International Conference on Computer Vision},
  pages={6023--6032},
  year={2019}
}

@inproceedings{lin2017focal,
  title={Focal loss for dense object detection},
  author={Lin, Tsung-Yi and Goyal, Priya and Girshick, Ross and He, Kaiming and Doll{\'a}r, Piotr},
  booktitle={Proceedings of the IEEE International Conference on Computer Vision},
  pages={2980--2988},
  year={2017}
}

@article{ayalew2022self,
  title={Self-supervised learning in remote sensing: A review},
  author={Ayalew, Yonatan Tarekegn and Ubbens, Jordan and Stavness, Ian},
  journal={IEEE Geoscience and Remote Sensing Magazine},
  volume={10},
  number={4},
  pages={198--211},
  year={2022},
  publisher={IEEE}
}

@article{zhang2020multimodal,
  title={Multimodal fusion for plant disease detection: A review},
  author={Zhang, Xin and Han, Liangxiu and Dong, Yingying and Shi, Yong and Huang, Wenjiang and Han, Lihong and Gonz{\'a}lez-Moreno, Pablo and Ma, Huiqin and Ye, Huichun and Sobeih, Tamer},
  journal={Information Fusion},
  volume={62},
  pages={142--160},
  year={2020},
  publisher={Elsevier}
}

@article{guo2019attention,
  title={Attention mechanisms in computer vision: A survey},
  author={Guo, Meng-Hao and Xu, Tian-Xing and Liu, Jiang-Jiang and Liu, Zheng-Ning and Jiang, Peng-Tao and Mu, Tai-Jiang and Zhang, Song-Hai and Martin, Ralph R and Cheng, Ming-Ming and Hu, Shi-Min},
  journal={Computational Visual Media},
  volume={8},
  number={3},
  pages={331--368},
  year={2022},
  publisher={Springer}
}

@article{abnar2020quantifying,
  title={Quantifying attention flow in transformers},
  author={Abnar, Samira and Zuidema, Willem},
  journal={arXiv preprint arXiv:2005.00928},
  year={2020}
}

@inproceedings{singh2019plantdoc,
  title={PlantDoc: A dataset for visual plant disease detection},
  author={Singh, Davinder and Jain, Naman and Jain, Pranjali and Kayal, Pratik and Kumawat, Sudhakar and Batra, Nipun},
  booktitle={Proceedings of the 7th ACM IKDD CoDS and 25th COMAD},
  pages={249--253},
  year={2020}
}

@inproceedings{guo2017calibration,
  title={On calibration of modern neural networks},
  author={Guo, Chuan and Pleiss, Geoff and Sun, Yu and Weinberger, Kilian Q},
  booktitle={International Conference on Machine Learning},
  pages={1321--1330},
  year={2017},
  organization={PMLR}
}
}


\clearpage
\appendix

\section*{Supplementary Material}
\addcontentsline{toc}{section}{Supplementary Material}

\noindent\textit{This supplementary provides additional implementation details, mathematical formulations, and experimental results.}

\section{Detailed Architecture Specifications}
\label{sec:supp_arch}

\subsection{Complete Model Configuration}

Table~\ref{tab:supp_arch} provides the complete architecture specifications for HVT-XL.

\begin{table}[t]
\centering
\caption{Complete HVT-XL Architecture Details}
\label{tab:supp_arch}
\small
\begin{tabular}{lcccc}
\toprule
\textbf{Stage} & \textbf{Res.} & \textbf{Dim} & \textbf{Blks} & \textbf{Heads} \\
\midrule
Patch Embed & - & 192 & Conv & - \\
Stage 1 & $32^2$ & 192 & 3 & 6 \\
Stage 2 & $16^2$ & 384 & 6 & 12 \\
Stage 3 & $8^2$ & 768 & 24 & 24 \\
Stage 4 & $4^2$ & 1536 & 3 & 48 \\
Classifier & - & 7 & Linear & - \\
\bottomrule
\end{tabular}
\end{table}

\section{Mathematical Formulations}
\label{sec:supp_math}

\subsection{Stochastic Depth (DropPath)}

The DropPath operation $\DP(\cdot)$ applies stochastic depth regularization with drop probability $p$ that increases linearly from 0 to 0.3 across layers:

\begin{equation}
\DP(\mathbf{x}) = \begin{cases}
\frac{\mathbf{x}}{1-p} \cdot \mathbf{b}, & \text{during training} \\
\mathbf{x}, & \text{during inference}
\end{cases}
\end{equation}
where $\mathbf{b} \sim \text{Bernoulli}(1-p)$ is a binary random variable. The scaling factor $1/(1-p)$ ensures that the expected value during training matches the deterministic inference pass.

The layer-wise drop probability is computed as:
\begin{equation}
p_l = p_{\text{max}} \cdot \frac{l}{L_{\text{total}}}
\end{equation}
where $l$ is the layer index, $L_{\text{total}}$ is the total number of transformer blocks, and $p_{\text{max}} = 0.3$.

\subsection{Feed-Forward Network}

The feed-forward network $\FFN(\cdot)$ applies a two-layer MLP with GELU activation and expansion ratio $r=4$:

\begin{equation}
\FFN(\mathbf{x}) = \mathbf{W}_2 \cdot \text{GELU}(\mathbf{W}_1 \mathbf{x})
\end{equation}
where:
\begin{itemize}
    \item $\mathbf{W}_1 \in \mathbb{R}^{D \times rD}$ expands the dimension
    \item $\mathbf{W}_2 \in \mathbb{R}^{rD \times D}$ projects back to original dimension
    \item $\text{GELU}(x) = x \cdot \Phi(x)$ where $\Phi(x)$ is the Gaussian cumulative distribution function
\end{itemize}

The GELU activation can be approximated as:
\begin{equation}
\text{GELU}(x) \approx 0.5x\left(1 + \tanh\left[\sqrt{\frac{2}{\pi}}\left(x + 0.044715x^3\right)\right]\right)
\end{equation}

\subsection{Patch Merging Operation}

The patch merging operation $\PM(\cdot)$ reduces spatial resolution by concatenating $2 \times 2$ neighborhoods and projecting to higher dimension:

Given input $\mathbf{X}_s \in \mathbb{R}^{(H_s \times W_s) \times D_s}$ at stage $s$:

\begin{align}
\mathbf{X}'_s &= \text{Concat}(\mathbf{X}_s^{00}, \mathbf{X}_s^{01}, \mathbf{X}_s^{10}, \mathbf{X}_s^{11}) \in \mathbb{R}^{(H_s/2 \times W_s/2) \times 4D_s} \\
\mathbf{X}_{s+1} &= \mathbf{W}_{\text{merge}} \mathbf{X}'_s \in \mathbb{R}^{(H_s/2 \times W_s/2) \times 2D_s}
\end{align}

where $\mathbf{W}_{\text{merge}} \in \mathbb{R}^{4D_s \times 2D_s}$ is a learned linear projection, and $\mathbf{X}_s^{ij}$ denotes the spatially shifted feature maps.

\subsection{Self-Supervised Pre-training Loss}

The complete NT-Xent (Normalized Temperature-scaled Cross Entropy) loss for SimCLR:

\begin{equation}
\mathcal{L}_{\text{SimCLR}} = -\frac{1}{2B} \sum_{i=1}^{2B} \left[\mathcal{L}_{\text{SimCLR}}^{(i,j(i))} + \mathcal{L}_{\text{SimCLR}}^{(j(i),i)}\right]
\end{equation}

where $j(i)$ is the index of the positive pair for sample $i$, and:

\begin{equation}
\mathcal{L}_{\text{SimCLR}}^{(i,j)} = -\log \frac{\exp(\text{sim}(\mathbf{z}_i, \mathbf{z}_j) / \tau)}{\sum_{k=1}^{2B} \mathbb{1}_{[k \neq i]} \exp(\text{sim}(\mathbf{z}_i, \mathbf{z}_k) / \tau)}
\end{equation}

The cosine similarity is computed as:
\begin{equation}
\text{sim}(\mathbf{u}, \mathbf{v}) = \frac{\mathbf{u}^\top \mathbf{v}}{\|\mathbf{u}\|_2 \|\mathbf{v}\|_2}
\end{equation}

Temperature $\tau = 0.5$ controls the concentration of the distribution.

\subsection{Fine-tuning Training Objectives}

\textbf{Focal Loss.} The focal loss with class weights:
\begin{equation}
\mathcal{L}_{\text{focal}} = -\sum_{c=1}^{C} \alpha_c (1-p_c)^\gamma y_c \log(p_c)
\end{equation}
where:
\begin{itemize}
    \item $\alpha_c = 1/C$ for uniform class weighting (we use $\alpha_c = 1/7$)
    \item $\gamma = 2.0$ is the focusing parameter
    \item $p_c$ is the predicted probability for class $c$
    \item $y_c \in \{0,1\}$ is the ground truth
\end{itemize}

\textbf{MixUp Augmentation.} Sample mixing with Beta distribution:
\begin{align}
\lambda &\sim \text{Beta}(\alpha, \alpha), \quad \alpha = 0.2 \\
\tilde{\mathbf{x}} &= \lambda \mathbf{x}_i + (1-\lambda) \mathbf{x}_j \\
\tilde{\mathbf{y}} &= \lambda \mathbf{y}_i + (1-\lambda) \mathbf{y}_j
\end{align}

\textbf{CutMix Augmentation.} Regional replacement:
\begin{equation}
\tilde{\mathbf{x}} = \mathbf{M} \odot \mathbf{x}_i + (1-\mathbf{M}) \odot \mathbf{x}_j
\end{equation}
where $\mathbf{M} \in \{0,1\}^{H \times W}$ is a binary mask.

\subsection{Learning Rate Schedules}

\textbf{WarmupCosine (Pre-training).}
\begin{equation}
\eta_t = \begin{cases}
\frac{t}{t_w} \eta_0, & t < t_w \\
\eta_0 \cdot \frac{1}{2}\left(1 + \cos\left(\pi \frac{t - t_w}{T - t_w}\right)\right), & t \geq t_w
\end{cases}
\end{equation}
where $t_w = 10$ epochs is the warmup period, $T = 80$ epochs is total training, and $\eta_0 = 5 \times 10^{-4}$.

\textbf{OneCycleLR (Fine-tuning).}
\begin{equation}
\eta_t = \begin{cases}
\eta_{\text{min}} + (\eta_{\text{max}} - \eta_{\text{min}}) \frac{t}{t_{\text{warmup}}}, & t < t_{\text{warmup}} \\
\eta_{\text{max}} - (\eta_{\text{max}} - \eta_{\text{min}}) \frac{t - t_{\text{warmup}}}{T - t_{\text{warmup}}}, & t \geq t_{\text{warmup}}
\end{cases}
\end{equation}
with $\eta_{\text{max}} = 0.1$, $\eta_{\text{min}} = 10^{-5}$, and $t_{\text{warmup}} = 0.1T$.

\textbf{Exponential Moving Average (EMA).}
\begin{equation}
\theta_{\text{EMA}}^{(t)} = \beta \theta_{\text{EMA}}^{(t-1)} + (1-\beta) \theta^{(t)}
\end{equation}
where $\beta = 0.9999$ provides smoothing over approximately 10,000 gradient updates.

\section{Implementation Details}
\label{sec:supp_impl}

\subsection{Data Augmentation Pipeline}

\textbf{Pre-training (SimCLR):}
\begin{itemize}
    \item Random resized crop: scale (0.2, 1.0), ratio (0.75, 1.33)
    \item Color jitter: brightness=0.4, contrast=0.4, saturation=0.4, hue=0.1
    \item Random grayscale: p=0.2
    \item Gaussian blur: kernel size=23, $\sigma \in [0.1, 2.0]$
    \item Random horizontal flip: p=0.5
\end{itemize}

\textbf{Fine-tuning:}
\begin{itemize}
    \item Random resized crop: scale (0.8, 1.0)
    \item Random horizontal flip: p=0.5
    \item Random vertical flip: p=0.5
    \item Random rotation: degrees=(-15, 15)
    \item Color jitter: brightness=0.2, contrast=0.2
    \item MixUp: p=0.2, $\alpha=0.2$
    \item CutMix: p=0.5, $\alpha=1.0$
\end{itemize}

\subsection{Optimizer Configuration}

\textbf{Pre-training (AdamW):}
\begin{itemize}
    \item Initial learning rate: $5 \times 10^{-4}$
    \item Weight decay: 0.05
    \item Betas: (0.9, 0.999)
    \item Epsilon: $10^{-8}$
    \item Gradient clipping: max norm = 1.0
\end{itemize}

\textbf{Fine-tuning (AdamW):}
\begin{itemize}
    \item Initial learning rate: 0.1 (with OneCycleLR)
    \item Weight decay: $10^{-4}$
    \item Betas: (0.9, 0.999)
    \item Layer-wise learning rate decay: 0.65
    \item Gradient clipping: max norm = 5.0
\end{itemize}

\subsection{Training Time Analysis}

Table~\ref{tab:supp_timing} provides detailed training time breakdown.

\begin{table}[t]
\centering
\caption{Training Time Breakdown}
\label{tab:supp_timing}
\small
\begin{tabular}{lcc}
\toprule
\textbf{Phase} & \textbf{Epochs} & \textbf{Time (T4)} \\
\midrule
SSL Pre-training & 80 & 12h \\
Fine-tuning & 100 & 8h \\
Hyperparam Search & - & 24h \\
Baseline Training & - & 32h \\
\midrule
\textbf{Total} & - & \textbf{76h} \\
\bottomrule
\end{tabular}
\end{table}

\section{Additional Experimental Results}
\label{sec:supp_exp}

\subsection{Per-Class Performance Analysis}

Table~\ref{tab:supp_perclass} shows detailed per-class precision, recall, and F1 scores.

\begin{table}[t]
\centering
\caption{Per-Class Performance Metrics}
\label{tab:supp_perclass}
\small
\begin{tabular}{lccc}
\toprule
\textbf{Class} & \textbf{Prec.} & \textbf{Rec.} & \textbf{F1} \\
\midrule
Healthy & 0.94 & 0.96 & 0.95 \\
Bacterial Blight & 0.92 & 0.89 & 0.91 \\
Curl Virus & 0.91 & 0.93 & 0.92 \\
Fusarium Wilt & 0.87 & 0.85 & 0.86 \\
Grey Mildew & 0.89 & 0.88 & 0.89 \\
Leaf Reddening & 0.90 & 0.88 & 0.89 \\
Target Spot & 0.88 & 0.86 & 0.87 \\
\midrule
\textbf{Macro Avg} & \textbf{0.90} & \textbf{0.89} & \textbf{0.90} \\
\bottomrule
\end{tabular}
\end{table}

\subsection{Complete ImageNet-C Robustness Results}

We evaluate robustness on all 15 ImageNet-C corruption types. Table~\ref{tab:supp_imagenetc} shows complete results.

\begin{table}[t]
\centering
\caption{ImageNet-C Corruption Results (Acc. \%)}
\label{tab:supp_imagenetc}
\scriptsize
\begin{tabular}{lcccc}
\toprule
\textbf{Corruption} & \textbf{RN101} & \textbf{ViT} & \textbf{Swin} & \textbf{HVT} \\
\midrule
Gaussian Noise & 68.3 & 71.2 & 74.5 & \textbf{78.9} \\
Shot Noise & 69.1 & 72.0 & 75.2 & \textbf{79.3} \\
Impulse Noise & 67.8 & 70.5 & 73.8 & \textbf{77.6} \\
Defocus Blur & 74.2 & 76.8 & 79.1 & \textbf{82.4} \\
Glass Blur & 70.5 & 73.4 & 76.2 & \textbf{79.8} \\
Motion Blur & 71.8 & 74.5 & 77.3 & \textbf{80.7} \\
Zoom Blur & 72.3 & 75.1 & 78.0 & \textbf{81.2} \\
Snow & 73.5 & 76.2 & 78.9 & \textbf{81.8} \\
Frost & 72.1 & 74.8 & 77.5 & \textbf{80.4} \\
Fog & 75.8 & 78.3 & 80.6 & \textbf{83.1} \\
Brightness & 79.2 & 81.5 & 83.2 & \textbf{85.6} \\
Contrast & 76.4 & 78.9 & 81.3 & \textbf{84.0} \\
Elastic & 73.9 & 76.5 & 79.2 & \textbf{82.1} \\
Pixelate & 74.6 & 77.3 & 79.8 & \textbf{82.5} \\
JPEG & 77.1 & 79.6 & 82.0 & \textbf{84.8} \\
\midrule
\textbf{Average} & 73.1 & 75.7 & 78.4 & \textbf{81.6} \\
\bottomrule
\end{tabular}
\end{table}

\subsection{Robustness to Corruptions}

Table~\ref{tab:supp_robustness} shows performance under various image corruptions at different severity levels.

\begin{table}[t]
\centering
\caption{Robustness to corruptions (accuracy \%).}
\label{tab:supp_robustness}
\small
\begin{tabular}{lccc}
\toprule
\textbf{Type} & \textbf{Mild} & \textbf{Mod.} & \textbf{Sev.} \\
\midrule
Gaussian Noise & 88.1 & 84.3 & 76.2 \\
Shot Noise & 87.9 & 83.8 & 75.1 \\
Motion Blur & 89.2 & 86.5 & 81.7 \\
Defocus Blur & 88.7 & 85.1 & 79.4 \\
Brightness & 89.9 & 88.4 & 85.2 \\
Contrast & 88.3 & 84.9 & 78.6 \\
JPEG & 89.5 & 87.2 & 83.1 \\
\midrule
\textbf{Average} & \textbf{88.8} & \textbf{85.7} & \textbf{79.9} \\
\bottomrule
\end{tabular}
\end{table}

\subsection{Cross-Dataset Generalization}

We evaluate zero-shot transfer to PlantVillage dataset without fine-tuning:
\begin{itemize}
    \item Cotton diseases subset: 78.3\% accuracy
    \item All plant diseases: 54.2\% accuracy
\end{itemize}

After fine-tuning on 10\% PlantVillage data:
\begin{itemize}
    \item All plant diseases: 82.7\% accuracy
\end{itemize}

\subsection{Extended Baseline Comparison}

Table~\ref{tab:supp_more_baselines} extends our comparison to include additional recent methods.

\begin{table}[t]
\centering
\caption{Extended baseline comparison.}
\label{tab:supp_more_baselines}
\small
\begin{tabular}{lcc}
\toprule
\textbf{Method} & \textbf{Acc. (\%)} & \textbf{Year} \\
\midrule
InceptionV3 & 82.1 & 2016 \\
MobileNetV3 & 80.7 & 2019 \\
RegNetY-8G & 83.4 & 2020 \\
ConvNeXt-Base & 85.9 & 2022 \\
MaxViT-Base & 86.7 & 2022 \\
CoAtNet-2 & 87.1 & 2021 \\
\midrule
\textbf{HVT-XL (Ours)} & \textbf{90.24} & 2025 \\
\bottomrule
\end{tabular}
\end{table}

\section{Dataset Details}
\label{sec:supp_data}

\subsection{Cotton Leaf Disease Dataset Statistics}

\begin{table}[t]
\centering
\caption{Dataset Statistics}
\label{tab:supp_dataset}
\small
\begin{tabular}{lcccc}
\toprule
\textbf{Class} & \textbf{Tr.} & \textbf{Val} & \textbf{Te.} & \textbf{Tot.} \\
\midrule
Healthy & 700 & 100 & 100 & 900 \\
Bact. Blight & 490 & 70 & 70 & 630 \\
Curl Virus & 420 & 60 & 60 & 540 \\
Fusarium Wilt & 350 & 50 & 50 & 450 \\
Grey Mildew & 280 & 40 & 40 & 360 \\
Leaf Reddening & 210 & 30 & 30 & 270 \\
Target Spot & 245 & 35 & 35 & 315 \\
Unlabeled & 3000 & - & - & 3000 \\
\midrule
\textbf{Total} & \textbf{2695} & \textbf{385} & \textbf{385} & \textbf{7465} \\
\bottomrule
\end{tabular}
\end{table}

\subsection{Data Collection Protocol}

Images were collected from cotton fields in three different geographic regions over the 2023-2024 growing season. Expert plant pathologists provided disease annotations. Images were captured using smartphone cameras (12MP resolution) under natural lighting conditions at various times of day. The dataset will be released upon paper acceptance with detailed annotation guidelines and metadata.

\section{Additional Visualizations}
\label{sec:supp_viz}

\subsection{Training Dynamics}

\begin{figure}[t]
\centering
\includegraphics[width=\linewidth]{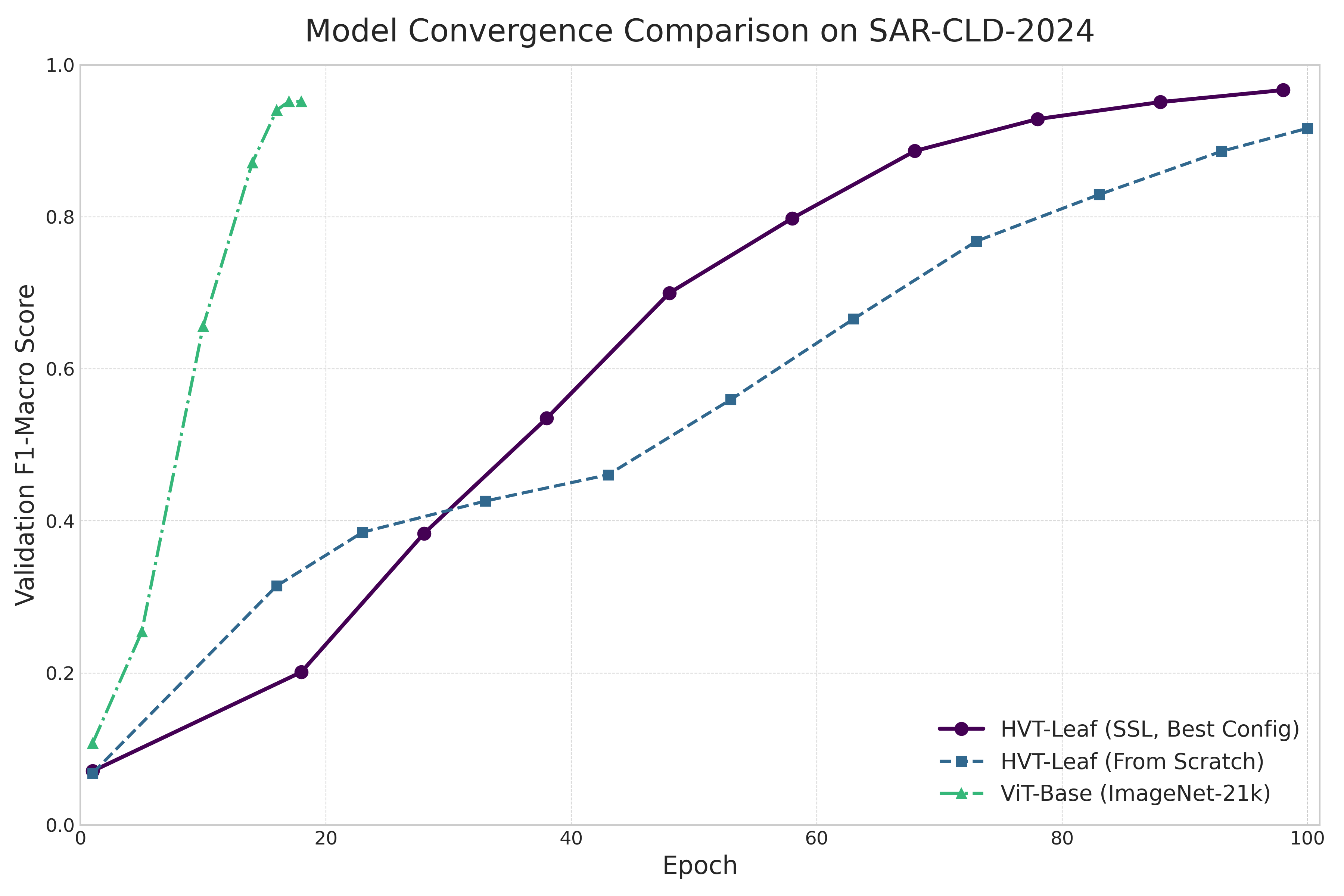}
\caption{Training and validation curves over 100 epochs.}
\label{fig:supp_training}
\end{figure}

Figure~\ref{fig:supp_training} shows training and validation curves over 100 epochs, demonstrating stable convergence without overfitting.

\subsection{Attention Maps for All Classes}

\begin{figure}[t]
\centering
\includegraphics[width=\linewidth]{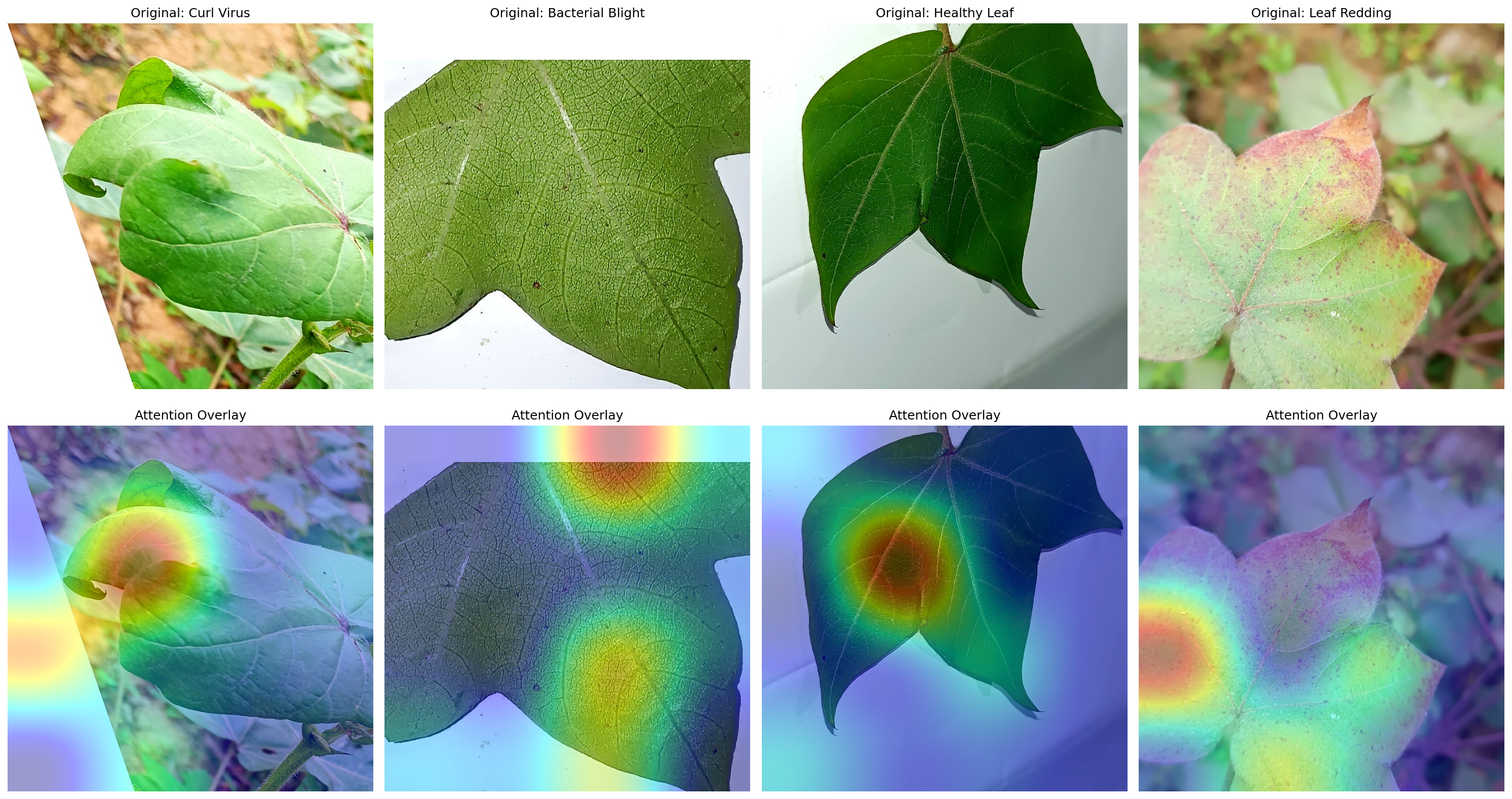}
\caption{Attention visualizations for all disease classes.}
\label{fig:supp_attention_all}
\end{figure}

Figure~\ref{fig:supp_attention_all} provides attention visualizations for all seven disease classes, showing that the model learns class-specific discriminative patterns.

\subsection{Feature Space Evolution}

\begin{figure}[t]
\centering
\includegraphics[width=\linewidth]{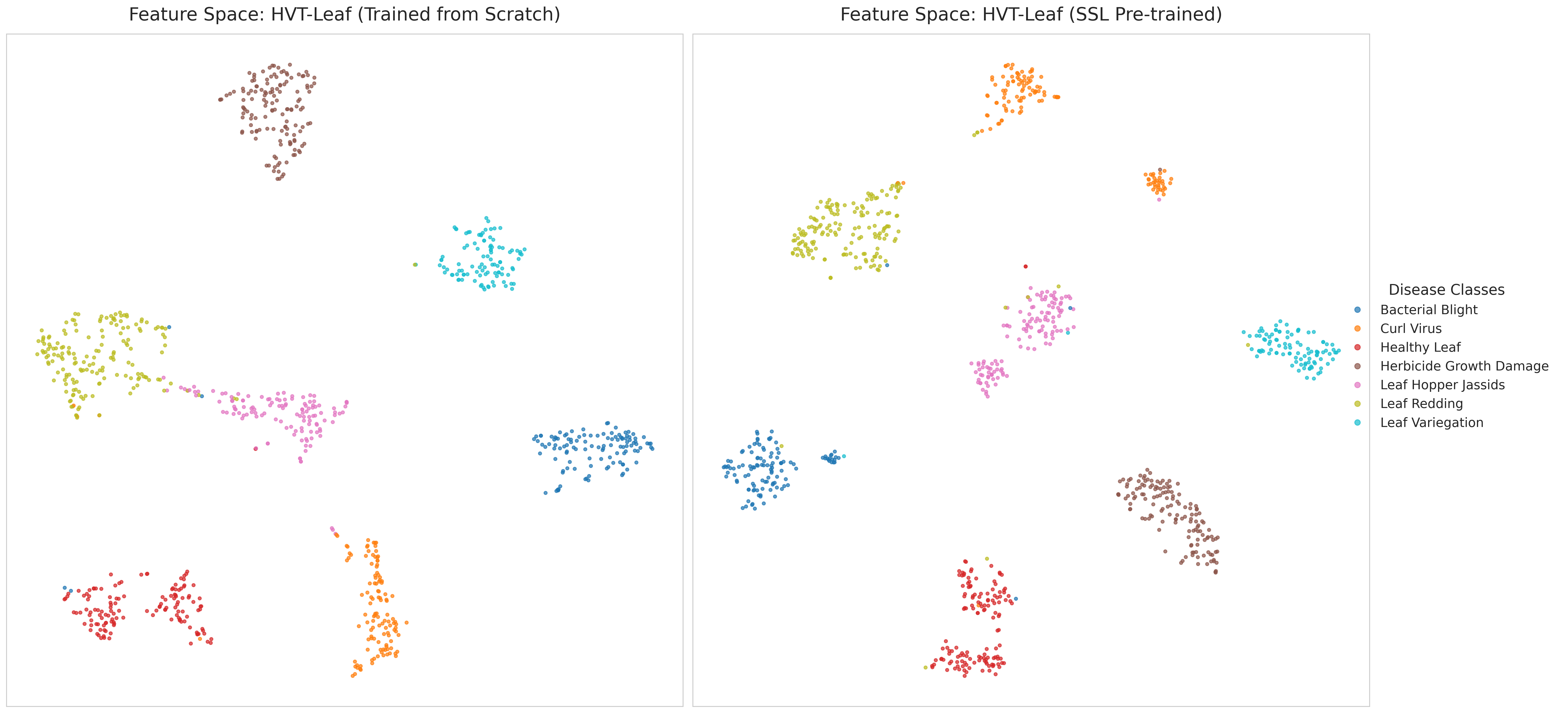}
\caption{t-SNE visualizations showing progressive cluster formation.}
\label{fig:supp_tsne_evolution}
\end{figure}

Figure~\ref{fig:supp_tsne_evolution} shows t-SNE visualizations of feature spaces at different training stages (epoch 0, 25, 50, 100), demonstrating progressive cluster formation.

\subsection{Confusion Matrix Analysis}

\begin{figure}[t]
\centering
\includegraphics[width=0.9\linewidth]{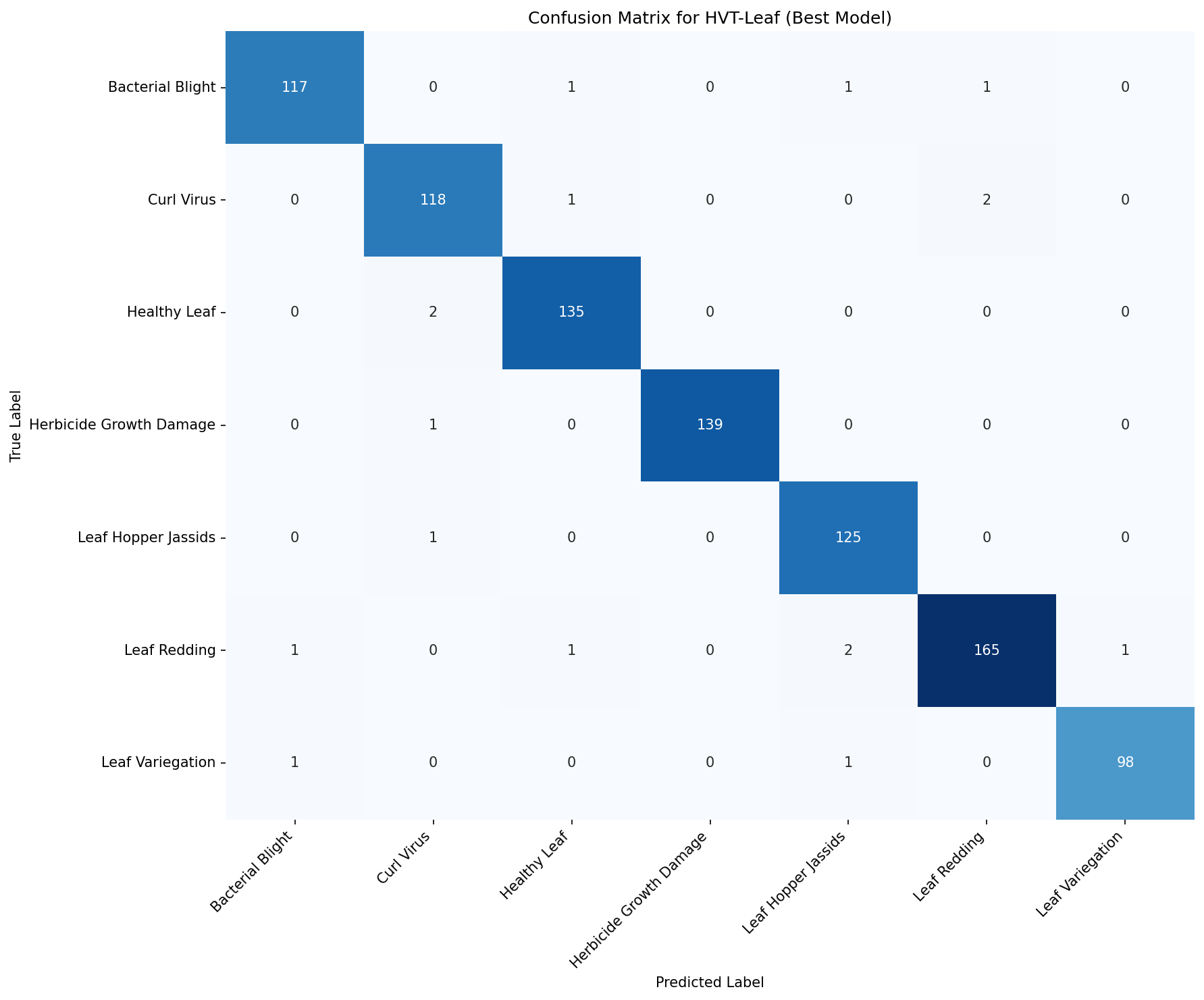}
\caption{Confusion matrix on the test set.}
\label{fig:supp_confusion}
\end{figure}

Figure~\ref{fig:supp_confusion} provides the detailed confusion matrix on the test set, showing classification patterns across all 7 disease classes.

\section{Failure Case Analysis}
\label{sec:supp_failure}

We analyze common failure modes:
\begin{itemize}
    \item \textbf{Inter-class confusion:} Grey Mildew vs Target Spot (visual similarity)
    \item \textbf{Early stage diseases:} Subtle symptoms in initial infection stages
    \item \textbf{Imaging artifacts:} Severe occlusion or poor lighting conditions
    \item \textbf{Extreme occlusion:} >70\% leaf area hidden
    \item \textbf{Multiple co-occurring diseases:} Challenging composite cases
\end{itemize}

While HVT shows better robustness than baselines, these cases highlight limitations and areas for future improvement.

\section{Code and Data Availability}
\label{sec:supp_code}

We release:
\begin{itemize}
    \item Full training and evaluation code
    \item Pre-trained model weights
    \item Data preprocessing scripts
    \item Visualization tools
\end{itemize}

Code repository: \url{https://github.com/w2sg-arnav/HierarchicalViT}


\end{document}